\documentclass[letterpaper]{article} 
\usepackage{aaai25}  
\usepackage{tabularx}
\usepackage{amsmath}
\usepackage{booktabs}
\usepackage{multirow}
\usepackage{times}  
\usepackage{helvet}  
\usepackage{courier}  
\usepackage[hyphens]{url}  
\usepackage{graphicx} 
\usepackage{natbib}  
\usepackage{caption} 
\usepackage{algorithm}
\usepackage{algorithmic}

\usepackage{newfloat}
\usepackage{listings}
\DeclareCaptionStyle{ruled}{labelfont=normalfont,labelsep=colon,strut=off} 
\floatstyle{ruled}
\newfloat{listing}{tb}{lst}{}
\floatname{listing}{Listing}
\title{An Evaluation Framework for Product Images Background Inpainting based on Human Feedback and Product Consistency}
\author{
    Yuqi Liang\textsuperscript{\rm 1 \rm2}, Jun Luo\textsuperscript{\rm 1}, Xiaoxi Guo\textsuperscript{\rm 1},
    Jianqi Bi\textsuperscript{\rm 1}\thanks{Corresponding author.}\\
}
\affiliations{
    \textsuperscript{\rm 1}Ant Group\\
    \textsuperscript{\rm 2}South China University of technology\\

    \{liangyuqi.lyq, wuguo.lj, xiaoxi.gxx, bijianqi.bjq\}@antgroup.com
}

\usepackage{bibentry}

\begin{document}

\maketitle

\begin{abstract}
In product advertising applications, the automated inpainting of backgrounds utilizing AI techniques in product images has emerged as a significant task. However, the techniques still suffer from issues such as inappropriate background and inconsistent product in generated product images, and existing approaches for evaluating the quality of generated product images are mostly inconsistent with human feedback causing the evaluation for this task to depend on manual annotation. To relieve the issues above, this paper proposes \emph{Human Feedback and Product Consistency (HFPC)}, which can automatically assess the generated product images based on two modules. Firstly, to solve inappropriate backgrounds, human feedback on 44,000 automated inpainting product images is collected to train a reward model based on multi-modal features extracted from BLIP and comparative learning. Secondly, to filter generated product images containing inconsistent products, a fine-tuned segmentation model is employed to segment the product of the original and generated product images and then compare the differences between the above two. Extensive experiments have demonstrated that HFPC can effectively evaluate the quality of generated product images and significantly reduce the expense of manual annotation. Moreover, HFPC achieves state-of-the-art(96.4\% in precision) in comparison to other open-source visual-quality-assessment models. Dataset and code are available at:
\url{https://github.com/created-Bi/background_inpainting_products_dataset/}.
\end{abstract}

%

\section{Introduction}

AI-based background inpainting techniques \citep{chen2024dnnam,verma2024graphfill,phutke2023image,feng2021deep,bhargavi2023zero,alam2024automated} have become crucial in improving the homogeneity of products and backgrounds in the visual needs of markets, notably in e-commerce and advertising, as well as sectors prioritizing high-quality images for profitability. \citep{wang2012country,hao2021two,he2021building}.

\begin{figure}[ht]
\centering
\includegraphics[width=\linewidth]{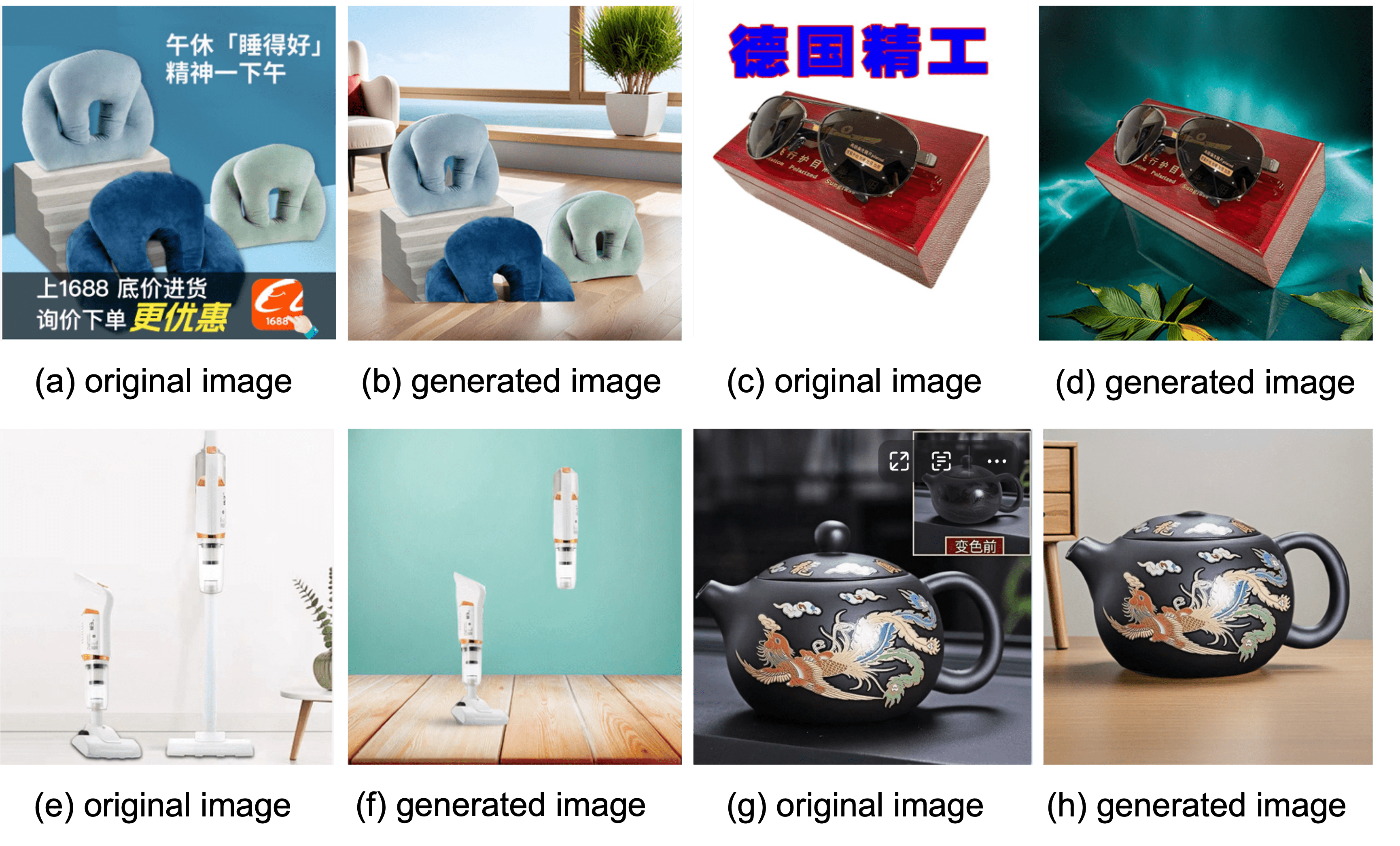}
\caption{Examples of original and AI-inpainting product images. Issues identified include (a-b) inappropriate background (pillows should not be placed on the ground), (c-d) inappropriate background(poor aesthetics), (e-f) product inconsistency between the original image and the generated image, (g-h) product inconsistency between the original image and the generated image.}
\label{fig:ori+gen}
\end{figure}

Compared to the rapid development of AI-based background inpainting techniques, the core challenge shifts to the evaluation of images generated by these techniques ~\citep{zhang2023image,qureshi2017critical,voronin2015no}. As illustrated in Fig. \ref{fig:ori+gen}, generated product images often suffer from the following problems: inappropriate background and product inconsistency. Currently, in real-world scenarios, a necessary way to filter these low-quality generated product images is human annotation, which is expensive and demanding.

The existing methods used for image quality assessment tend to evaluate generated images in isolation, without considering the need to preserve the integrity of the original product, position it appropriately, or ensure its harmony with the background. 
Common automatic evaluation metrics, such as the referenced metrics Fréchet Inception Distance(FID) \citep{yu2021frechet} and  Cumulative Maximum Mean Discrepancy(CMMD) \citep{jayasumana2024rethinking}, focus on image distribution but usually fail to capture nuances in individual images. Reference-free metrics such as CLIP-IQA \citep{wang2023exploring} merely assess the appearance and aesthetics of the images. More recent approaches exert to collecting human preferences and train reward models to predict the preferences for improving the generative models\citep{kirstain2023pick}. 
However, there is a gap between the definition of an excellent image in these methods and that in the background inpainting field(i.e., human feedback).

To relieve these issues, a novel evaluation framework named Human Feedback and Product Consistency(HFPC) was proposed to evaluate the quality of background inpainting product images. Firstly, to solve inappropriate backgrounds, human feedback on 44,000 automated inpainting product images is collected to train an image-referenced reward model. Specifically, in the training process, two sets of image pairs such as the (original, good) and (original, bad) generated images were constructed.
Secondly, to filter generated product images that contain inconsistent products, a fine-tuned segmentation model is used to segment the products of the original and generated product images\citep{liu2023grounding,xiong2024efficientsam} and then compare the differences between the above two. By visually comparing the differences between the segmentation masks, whether the product is preserved in its entirety can be determined.


Experimentally, HFPC achieves state-of-the-art (SOTA) performance on our HFPC-44k dataset compared to existing image-quality evaluation models. In addition, our product consistency evaluator effectively identifies issues such as product loss during image transformation.

The main contributions of this paper are as follows:
\begin{itemize}
    \item Introduce the HFPC-44k dataset, identifying two major issues in AI-based background inpainting images: inappropriate backgrounds or inconsistent products.
    \item Propose HFPC, a novel framework specifically designed for the automatic quality assessment of AI-based background inpainting images.
    \item Extensive experiments demonstrate that HFPC achieves high accuracy in filtering out the generated product images with inappropriate backgrounds or inconsistent products, significantly reducing the expense of manual annotation.
\end{itemize}

\section{Related Work}
\subsection{AI-based Background Inpainting for product Images}
AI-based background inpainting of product images involves several key techniques, including target segmentation, background retrieval, and image fusion\citep{chen2024dnnam,verma2024graphfill,phutke2023image,feng2021deep,bhargavi2023zero,alam2024automated}. First, to obtain the products, target segmentation of the original image is required. Target segmentation not only detects the product but also generates its accurate mask. Next, a suitable background is retrieved in the background gallery. The background gallery can either be collected manually or generated by Generative Adversarial Networks (GANs), e.g. StyleGAN\citep{karras2019style} can generate high-quality and realistic background images. Finally, the product and background are fused using image fusion techniques. The Deep Image Blending(DIB) method\citep{zhang2020deep} optimizes the image fusion results by combining the loss of Poisson fusion, as well as the loss of style and content computed from the deep network, and iteratively updating the pixels using the L-BFGS\citep{moritz2016linearly} solver to achieve smoothing in the gradient domain and consistency of texture within the image blending region.

\subsection{Image-to-image Evaluation}
In the field of automatic image generation evaluation, predominant metrics such as the Inception Score (IS) \citep{barratt2018note}, FID \citep{yu2021frechet}, and CMMD \citep{jayasumana2024rethinking} assess image quality by comparing statistical properties between original and generated images. However, these reference-based metrics often fail to capture detailed discrepancies at the individual image level. Alternatively, reference-free metrics like CLIP-IQA \citep{wang2023exploring}, leveraging the visual-linguistic capabilities of the pretrained CLIP model \citep{radford2021learning}, aim to directly evaluate image quality perception and abstract representation. Despite their advances, these metrics struggle, particularly in evaluations involving product images, where they inadequately reflect the appropriateness between products and their backgrounds and fail to accurately detect morphological changes and in products. This indicates a significant challenge in using current metrics for the assessment of AI-based background inpainting product images.

\subsection{Text-to-image Reward Models}
Recently, there have been several new studies in the area of using human preference training reward models to evaluate text-generated image models. \citet{xu2024imagereward} created a human preference dataset by asking users to rank a series of images and score their quality. Based on this dataset, they trained a human preference learning reward model called ImageReward and proposed a new method of using the ImageReward model to optimize a diffusion model through reward feedback learning (ReFL). In addition, \citet{kirstain2023pick} constructed a web application, Pick-a-Pic, which collects human preferences by allowing users to pick the better one from a pair of generated images, and this application has collected more than half a million images from different T2I models (e.g., Stable Diffusion 2.1, Dreamlike Photoreal 2.05, and Stable Diffusion XL) ~\citep{podell2023sdxl,wu2024multimodal,gupta2024progressive}, and they used these human preference data to train a CLIP-based scoring function, called PickScore, for predicting human preferences. In addition, ~\citet{wu2023human} also collected a large dataset of images generated by human choice and used this data set to train a classifier that produces a Human Preference Score (HPS), and demonstrated that image generation quality can be significantly improved by tuning the Stable Diffusion model. More recently, Liang et al. collected a rich feedback dataset containing artifactual regions, misaligned regions, misaligned keywords, and four fine-grained scores, and trained a multimodal model(RAHF)\citep{liang2024rich} to automatically predict these feedbacks.

\section{Method}


\begin{figure*}[!htbp]
  \centering
  \captionsetup{skip=1pt}
  \includegraphics[width=0.8\textwidth]{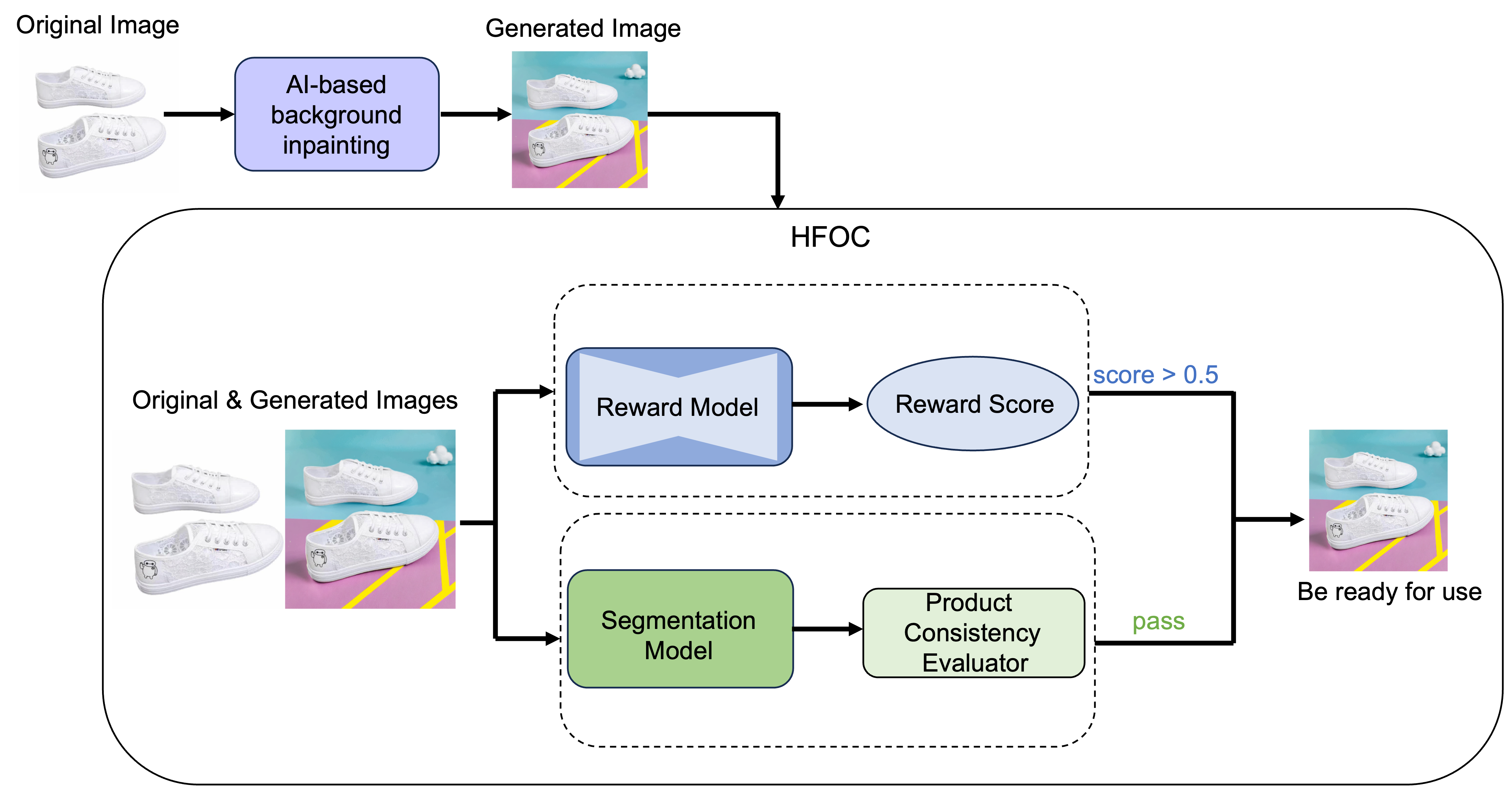}
  \caption{The HFPC contains two modules working in parallel. The first module is a reward model based on the multimodal BLIP, which scores a pair of original and generated images reflecting the appropriateness of the background. The second module is a product consistency assessment model.}
  \label{fig:intro}
\end{figure*}

\subsection{Overall Framework}


As shown in Fig. \ref{fig:intro}, this paper introduces the HFPC architecture, designed to automatically evaluate the quality of AI-based background inpainting product images. The HFPC architecture comprises two parallel modules. Module 1, a multimodal BLIP-based\citep{li2022blip} reward model, processes both the original and generated images to output a score reflecting the appropriateness of the background. Module 2, the product consistency assessment model, utilizes a fine-tuned segmentation model to segment products from the original and generated images. These segmented products are then compared using a product consistency evaluator. An image is deemed acceptable if it meets predefined thresholds for both background appropriateness and product consistency.

\subsection{Collection of HFPC-44k Dataset}

HFPC-44k comprises 44,000 pairs of original images and AI-based background inpainting product images, each annotated with corresponding human feedback. 
Let $D$ represent the entire dataset, where each element $d \in D$ consists of three components: : the original image $I$, a set of generated images $G = \{  G_{1}, \cdots, G_{i}, \cdots, G_{K} \}$, and corresponding label set $L = \{  L_{1}, \cdots, L_{i}, \cdots, L_{K} \} $. Here, $L_{i} = 1$ denotes that the human review of $i$-th generated image is passed and $L_{i} = 0$ means that the review is not passed.



\begin{figure*}[h]
    \centering
    \includegraphics[width=0.6\linewidth]{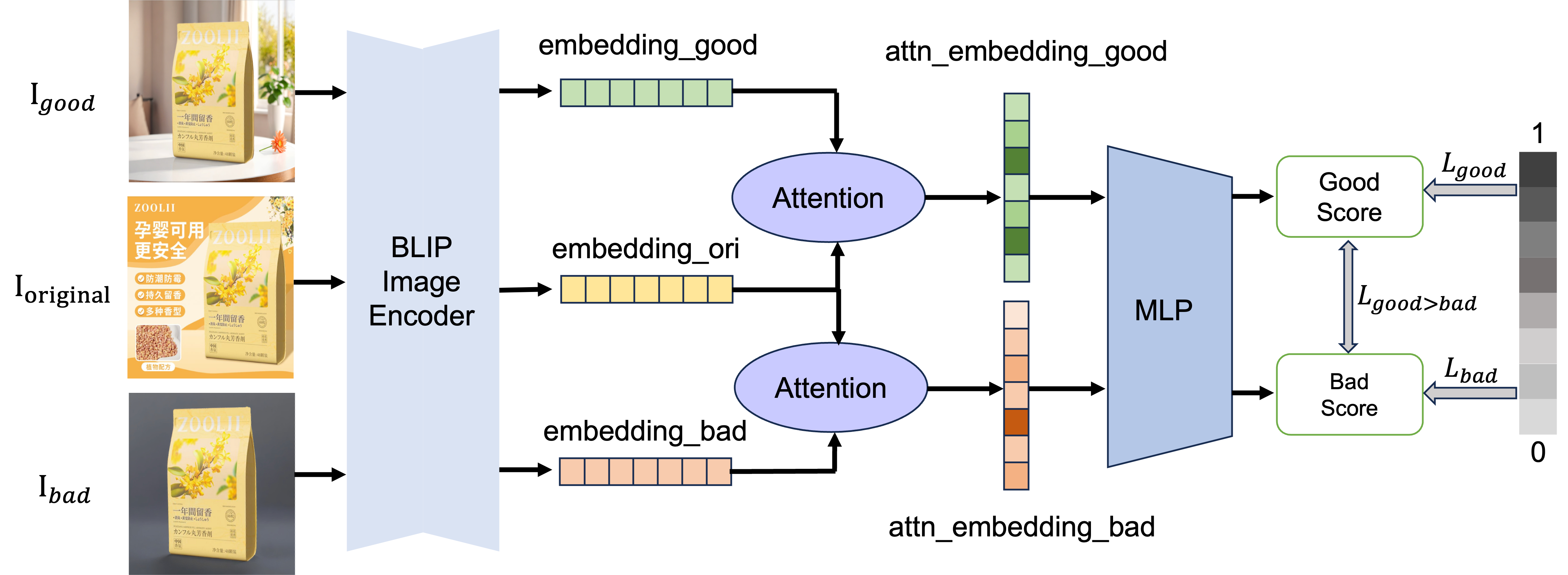}
    \caption{Reward model}
    \label{fig:rank model}
\end{figure*}

\subsection{Image-Referenced Reward Model}

Inspired by the success of the reward model in the textual field, an Image-Referenced Reward Model(IRRM) applying the image encoder from BLIP, which is shown in Fig. \ref{fig:rank model} was proposed. Firstly, IRRM receives an original image and two high-quality and low-quality AI-based background inpainting images. These images will be fed into the image encoder of BLIP to extract the respective image features. Subsequently, an attention module\citep{vaswani2017attention} is employed to enhance the perception of the original image features by comparative learning ~\citep{lei2016comparative}. Namely, IRRM combines the information of the original and generated product images by focusing on the key difference in features between high-quality and low-quality images. The attention mechanism not only improves IRRM's capacity to recognize low-quality images but also improves IRRM's interpretability. Secondly, the fused features are passed to a multi-layer perceptron(MLP) to obtain two categories of output. One of them is a high-quality image score and the other is low-quality. Finally, a triplet loss function is applied during the model training process to carefully optimize the model performance and ensure that the model can accurately differentiate between high-quality and low-quality images.

\subsection{Product Consistency Assessment Module}
\begin{figure*}[h]
    \centering
    \includegraphics[width=0.7\linewidth]{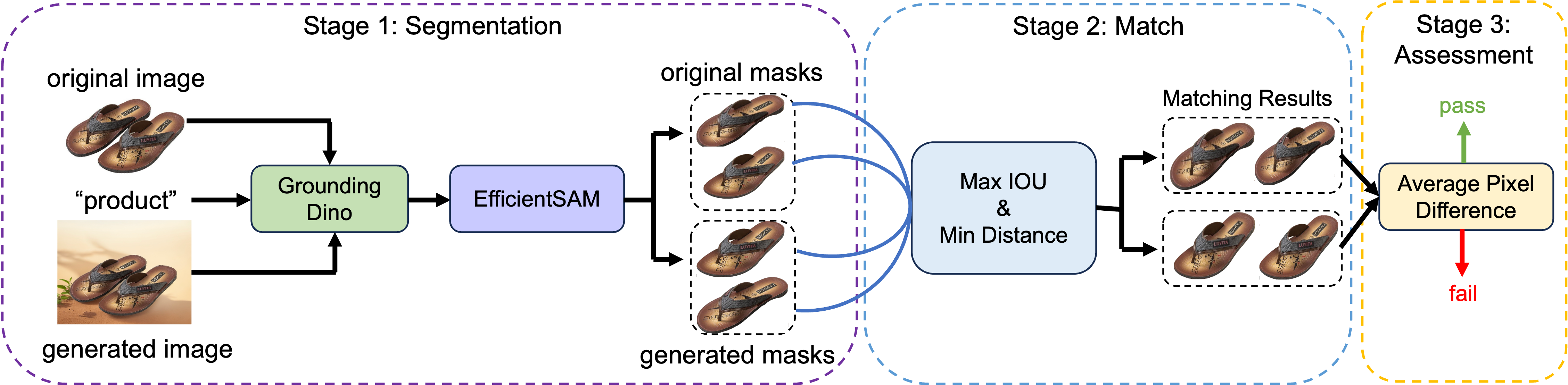}
    \caption{Product consistency assessment model}
    \label{fig:efficient-sam}
\end{figure*}

This module evaluates the consistency of products in original and generated images. Our method comprises three stages, as depicted in Figure \ref{fig:efficient-sam}.

First, both the original and generated images are inputted into a fine-tuned EfficientSAM model\citep{xiong2024efficientsam}. This model aims to obtain segmentation masks for all products in the images. To provide EfficientSAM with high-quality bounding box prompts, the Grounding Dino was integrated with EfficientSAM. The Grounding Dino model processes the input images using a text prompt "product." and then produces specific detection boxes for the products. 
Leveraging the detection boxes provided by Grounding Dino, the EfficientSAM model then generates segmentation masks for all products.
Moreover, more than 500 product images are manually annotated to fine-tune the segmentation model at the pixel level. The text prompt "products" were uniformly used during training to ensure consistent cueing.
Let the $m$ masks generated from the original image is denoted as $M_{ori} = \{  M_{1}, \cdots, M_{i}, \cdots, M_{m} \}$. $n$ masks generated from the generated image are denoted as $N_{gen} = \{  N_{1}, \cdots, N_{j}, \cdots, N_{n} \}$.

Given that the product's masks in $M_{ori}$ and $N_{gen}$ are unordered, it is necessary to match corresponding products between the original and generated images to accurately assess consistency. The matching is based on the shape and positional order of the products, under the assumption that the same products will exhibit similar shapes and positional arrangements in both images. For each mask $M_{i}$ in $M_{ori}$, the Intersection over Union (IoU) and positional distance with each mask $N_{j}$ in $N_{gen}$ was calculated, matching $M_{i}$ with the mask $N_{match}^i \in N_{gen}$ that has the highest IoU and smallest positional distance. This process results in a set of matched product pairs $MN = \{ (M_{1},N_{match}^1), \cdots\ (M_{i},N_{match}^i), \cdots (M_{m},N_{match}^m) \}$.

For each matched pair $(M_{i}, N_{j})$, consistency by computing the average pixel difference within the mask region of the products was then assessed. The difference serves as a metric for visual dissimilarity between the products (i.e., product inconsistency).

\subsection{Model Optimization of Reward Model}
For the reward model, it aims to evaluate the appropriateness of the background. The model is required to ensure that the high-quality images gain a higher score than the low-quality images. To achieve this goal, a composite loss function is designed to supervise the learning process.

Let $x_{i}$ be the original image, $y_i = [y_{i,1}, y_{i,0}]$ is the generated image obtained from the original image $x_{i}$ after AI-based background inpainting, where $y_{i,1}$ is the generated image that the human review passes, and $y_{i,0}$ is the generated image that the human review fails, and the corresponding target label is $t _i = [1, 0]$, where 1 means pass and 0 means fail. The $r_i = [r_{i,1}, r_{i,0}]$ denotes the model's predicted scores for the generated image $y_{i}$, where $r_{i,1}$ and $r_{i,0}$ represent the scores of the high-quality and low-quality images, respectively. Our overall loss function is defined as follows,

\begin{equation}
\mathcal{L}_{\text{total}} = \mathcal{L}_{\text{rank}} + \mathcal{L}_{\text{class}}
\end{equation}
, where $\mathcal{L}_{\text{rank}}$ is the ranking loss and $\mathcal{L}_{\text{class}}$ is the classification loss.

The ranking loss uses cross-entropy to encourage positive sample pairs to score higher than negative sample pairs,
\begin{equation}
\mathcal{L}_{\text{rank}} = \frac{1}{N} \sum_{i=1}^N \text{L}_{\text{CE}}(r_i, t_i)
\end{equation}
, where $\mathcal{L}_{\text{CE}}$ denotes the cross-entropy loss function ~\citep{mao2023cross}.

The classification loss is designed to ensure that the model prediction scores $\sigma(r_{i,1})$ greater than 0.5 for high-quality images and less than 0.5 for low-quality images. Classification loss is defined as follows, 
\begin{equation}
\mathcal{L}_{\text{class}} = \frac{1}{N} \sum_{i=1}^N (\text{L}_{BCE}(\sigma(r_{i,1}), 1) +   \text{L}_{BCE}(\sigma(r_ {i,0}), 0))
\end{equation}
, where $\sigma(\cdot)$ denotes the sigmoid function and $\mathcal{L}_\text{BCE}$ denotes the binary cross-entropy loss.
The relatively comprehensive loss function can simultaneously optimize the model's ranking ability and classification accuracy, thereby improving the overall performance of image quality assessment.

\section{Experiment}

In this section, how to expand the HFPC-44k dataset will be introduced. Considering the imbalance of different categories of images in real business scenarios (e.g., the number of photos in the apparel category is larger than that in the electrical appliances category), the training data is expanded accordingly. Moreover, the significant advantages of our approach are demonstrated in evaluating background inappropriateness and product consistency through extensive experiments.

\subsection{Expansion of Training Data}

HFPC-44k contains 44,244 pairs of original images of products and images generated by AI-based background inpainting. Each data is labeled with the generation time, the original image URL, the generated image URL, and the manual annotation tag (labeled as "passed" or "failed"). 35,000 pairs of images from this dataset were extracted for the model training process and the remaining 8,372 pairs of images were used in the test process. Among the 35,000 training samples, 80\% are split to build the training dataset, and 20\% are for the validation set.

\begin{figure}[h]
    \centering
    \includegraphics[width=\linewidth]{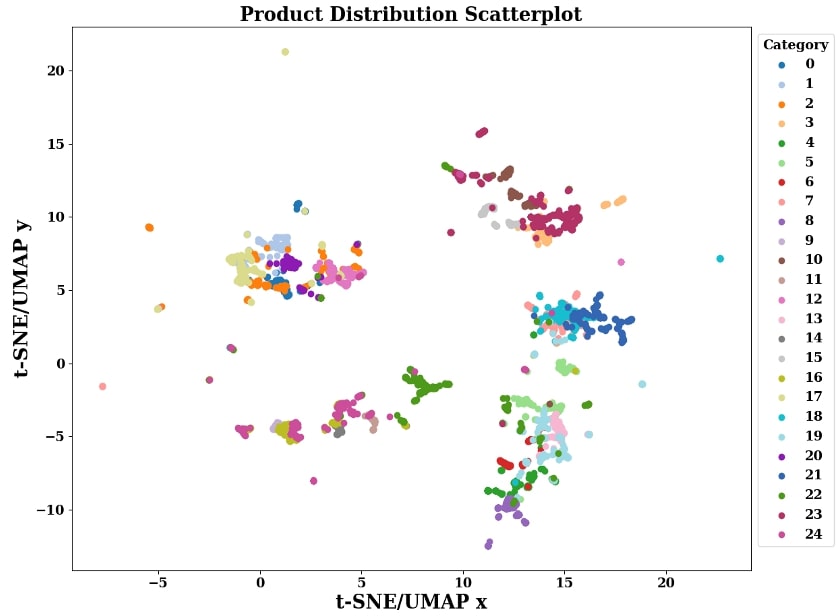}
    \caption{Visualization of clustering of product images. 
    The original image features were extracted from the BLIP encoder and clustered by KMeans and 25 categories were finally determined. Each color in the image represents one cluster, for example, the cluster with the highest number of products is the shoe category represented by blue, and the cluster with the lowest number is the cosmetics category (tubes) represented by grey.}
    \label{fig:cluster}
\end{figure}

\begin{figure}[h]
    \centering
    \includegraphics[width=\linewidth]{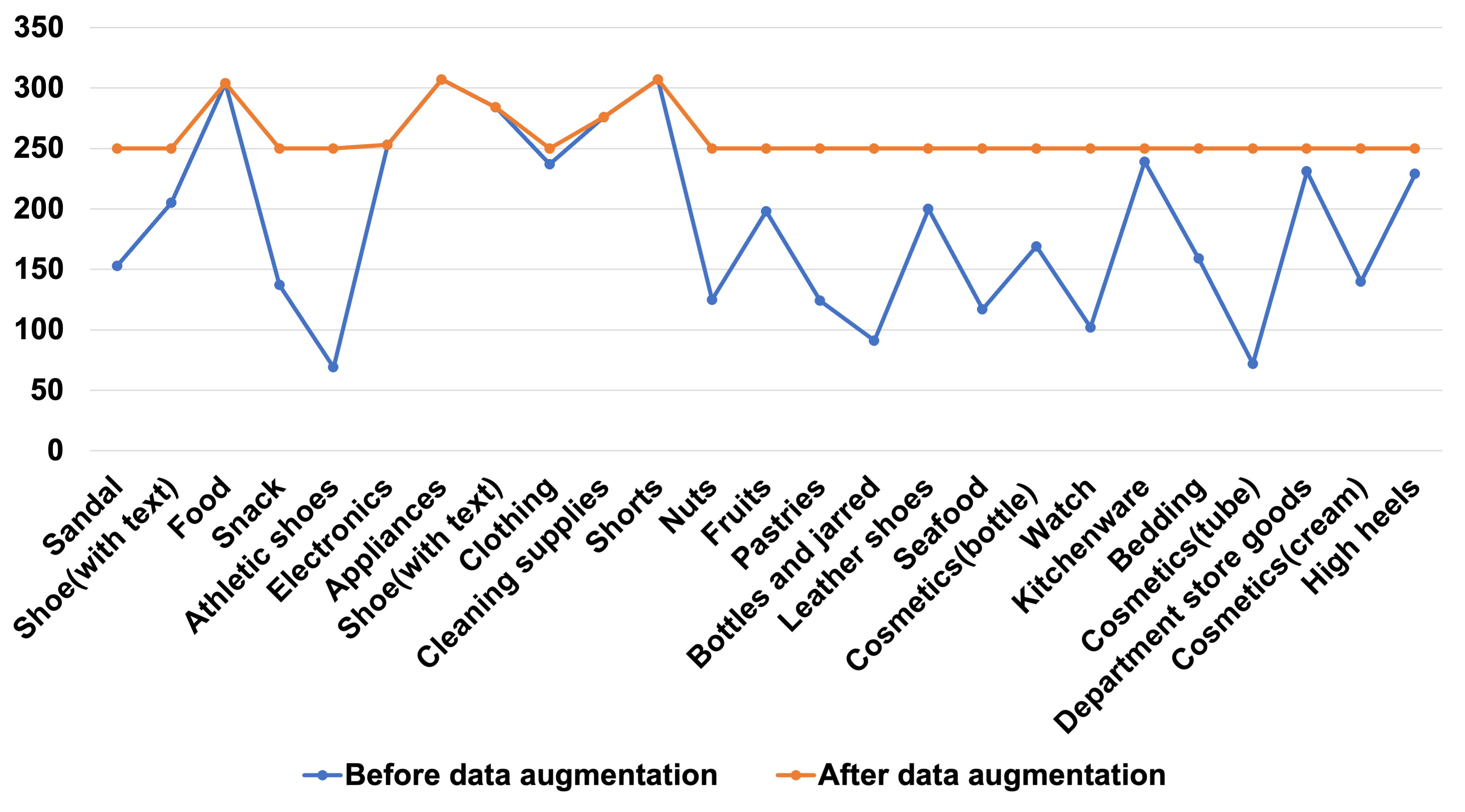}
    \caption{Comparison of product category distributions before and after augmentation. The image shows the change in the number of product images in each category before and after the optimization of the training dataset. Before data expansion, the number of product pictures in each category was not balanced, which is reflected in the inconsistently high and low blue lines. After data expansion, the orange line shows that the number of product images in all categories has reached a more balanced state.}
    \label{fig:product}
\end{figure}

It can be noticed that the number of images of certain commodity categories appeared to be extremely unbalanced. 
After extracting the features of all the original images of the commodities in the training set using a BLIP encoder, clustering by KMeans was performed to check the data distribution\citep{krishna1999genetic}. 
Finally, the most suitable number of clustering categories (i.e., number of clusters) is defined as 25. Furthermore, as shown in Fig. \ref{fig:cluster}, the number of product images corresponding to the blue points representing the shoes category is much higher than the red points representing the pots and pans category. The imbalance of data categories may cause the model to be biased in rating the product images of certain categories. 
Thus, a balanced method was applied to the training dataset. Specifically, a few numbers of product images are repeated using a random sampling of products in the same category.
Specific measures include data expansion for categories with a low number of images. As shown in Fig. \ref{fig:product}, the blue and orange lines represent the number of product images in each category before and after data expansion, respectively. In this way, the model can be accessed to a balanced number of different categories of product images during the training process.

\subsection{Evaluation Protocol}
Our HFPC model was trained based on 28,000 training samples and hyperparameters are tuned by performance on 7,000 validation samples. Finally, the performance of the model was validated on 8,000 test samples.

In the main background coordination scoring task, the Pearson linear correlation coefficient (PLCC) and the Spearman rank correlation coefficient (SRCC) ~\citep{sedgwick2014spearman} were reported. These two metrics are commonly used to evaluate scoring systems, measuring the accuracy and consistency of model performance in terms of linear and rank correlation, respectively ~\citep{talebi2018nima,zhai2020perceptual,yang2022maniqa}.

For the task of assessing the overall quality of images generated by AI-based background inpainting, including the assessment of background appropriateness and product consistency, three key metrics were considered as 1)low-quality image filtering accuracy rate (Pb), 2)low-quality image recall rate (Rb) and 3) high-quality image recall rate (Rg). The  calculation process of each metric is as follows,

\begin{enumerate}
    \item Precision rate for low-quality image filtering \(P_b\),
    \begin{equation}
    P_b = \frac{N_{\text{filtered, low-quality}}}{N_{\text{filtered}}}
    \end{equation}
    , where \(N_{\text{filtered, low-quality}}\) is the number of low-quality images actually filtered out and \(N_{\text{filtered}}\) is the total number of images filtered out.

    \item Recall rate for low-quality generated images \(R_b\),
    \begin{equation}
    R_b = \frac{N_{\text{filtered, low-quality}}}{N_{\text{original, low-quality}}}
    \end{equation}
    , where \(N_{\text{filtered, low-quality}}\) is the total number of filtered low-quality generated images and \(N_{\text{original, low-quality}}\) is the total number of original low-quality generated images.

    \item Recall rate for high-quality generated images \(R_g\),
    \begin{equation}
    R_g = \frac{N_{\text{filtered, high-quality}}}{N_{\text{original, high-quality}}}
    \end{equation}
    , where \(N_{\text{filtered, high-quality}}\) is the total number of filtered high-quality generated images and \(N_{\text{original, high-quality}}\) is the total number of original high-quality generated images.
    
\end{enumerate}

Measuring the above four key metrics can comprehensively provide strong support to evaluate the performance of the HFPC, even for further optimization.

\subsection{Baseline}
In this study, we evaluated a variety of image quality assessment (IQA) models to compare their performance with HFPC. Specifically, we assessed four popular QA metrics commonly used in the field of image generation. These metrics include both pretrained models, such as NIMA \citep{talebi2018nima}, and training-free approaches, including CLIP-Score \citep{hessel2021clipscore} and CLIP-IQA \citep{wang2023exploring}. Additionally, to ensure a fair comparison, we trained ImageReward\citep{xu2024imagereward}, a robust baseline in the domain of text-to-image generation IQA.

CLIP-Score: CLIP-Score utilizes the pre-trained CLIP model to extract the embedding vectors of the original image and the AI-based background inpainting image, and then calculates the cosine similarity of these two vectors to assess the consistency between the two images.The higher the CLIP-Score, the better the quality of the generated image.

CLIP-IQA: CLIP-IQA directly evaluates two aspects of the generated images with the help of the rich visual language capabilities of the pre-trained CLIP model: quality perception appearance, including exposure and appearance level; and abstract perception, i.e., emotion and aesthetics. The higher the score of CLIP-IQA, the better the quality of the generated images in terms of appearance and feeling. 

NIMA: NIMA is an image quality evaluation metric developed by Google that predicts the aesthetic and technical quality of an image through deep learning techniques. The system predicts human objective perception of an image by scoring it, and the output score is often used to reflect the overall quality or aesthetics of the image. The NIMA model is typically based on a trained Convolutional Neural Network (CNN), which learns evaluation patterns and criteria from a large number of scored images.

ImageReward: ImageReward is a universal human preference reward model for text-to-image generation, designed to effectively encode and reflect human preferences in image generation tasks. The model is trained through a systematic annotation process, including scoring and ranking, to learn and capture human subjective evaluations.

\subsection{Overall Results}

\begin{figure}[ht]
    \centering
    \includegraphics[width=\linewidth]{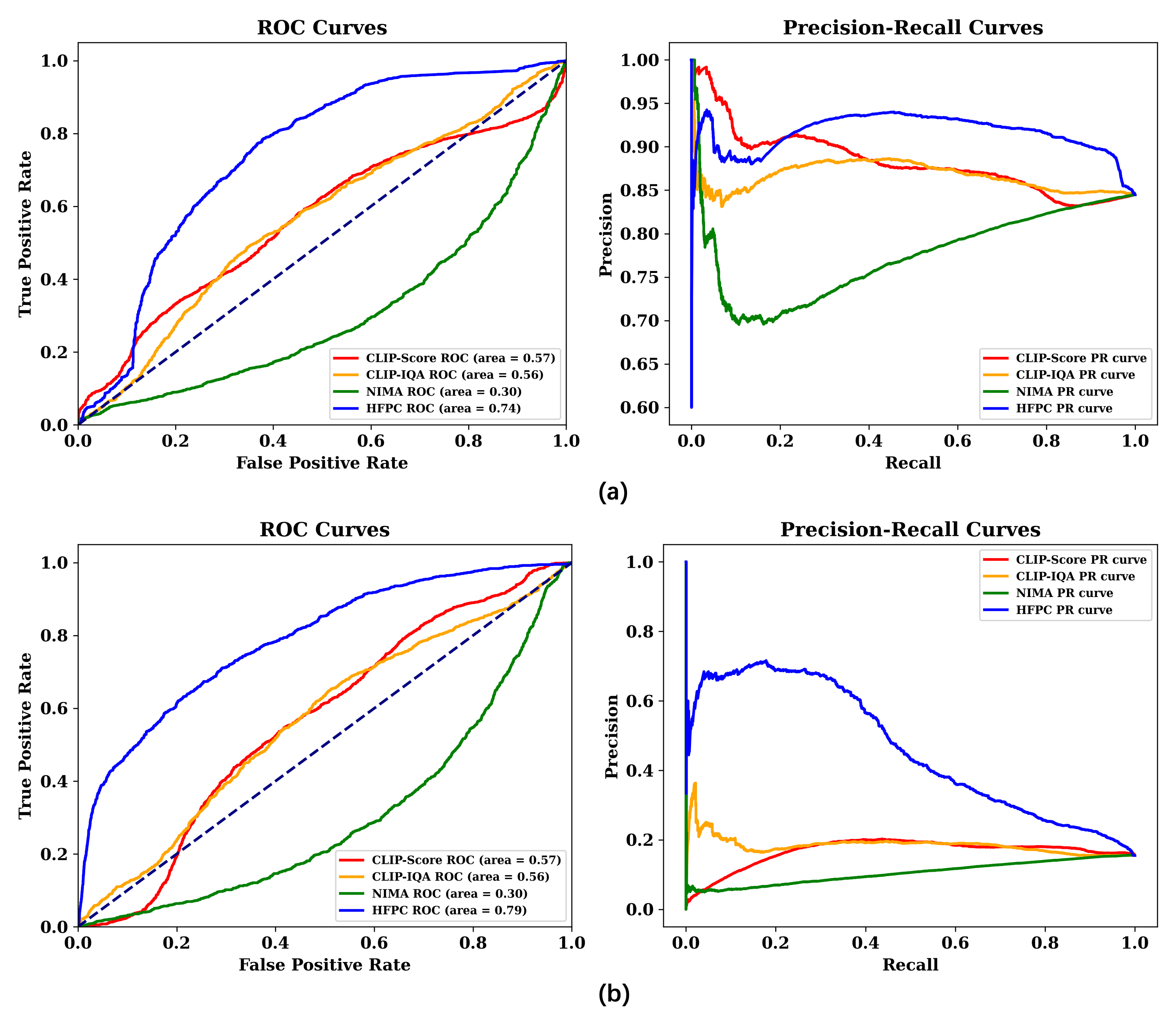}
    \caption{Showing the P-R curve and ROC curve for different methods. Figure (a) plots the ROC curves and P-R curves for low-quality images, focusing on the filtering ability of low-quality images. Figure (b) plots the ROC curves and P-R curves for high-quality images, focusing on the recall ability of high-quality images.}
    \label{fig:roc}
\end{figure}

The results of each method evaluation are shown in Table \ref{tab:assessment_results}. From the table, it can be seen that our designed HFPC framework performs the best in all indicators. In addition, the P-R and ROC curves of each method are plotted in Fig. \ref{fig:roc}. The two metrics can demonstrate the performance of each approach in the case of pass and fail. Based on these curves, the evaluation effects of different modules can be more intuitively compared, leading to further proof of the superiority of the HFPC framework in image quality evaluation.
In addition, the effects of the reward model and the product consistency assessment module in HFPC were verified from the ablation study, respectively. As shown in Table\ref{tab:assessment_results}, a single reward model already shows a performance very consistent with human preference, based on which the product consistency assessment module can further filter out the existence of product inconsistency images.

\subsection{Ablation Study}
To investigate the effect of each module on the performance of the model, we conducted addictional ablation experiments, including fusion of original and generated images via attention mechanisms, and classification and ranking losses. Particularly, we compared ImageReward-a ranking model without original image and with only ranking loss. As shown in Table\ref{tab:ablation study}, introducing original image data via attention mechanisms significantly improves model performance, and classification loss further enhances performance.

\begin{table}[ht]
\centering
\caption{Comparison of assessment results. \textsuperscript{+}indicates that the method requires training.}
\label{tab:assessment_results}

\renewcommand{\arraystretch}{1.5} 
\setlength{\tabcolsep}{10pt} 
\resizebox{\columnwidth}{!}{ 
\begin{tabular}{@{}lccccl@{}} 
\toprule 
\textbf{Method} & \textbf{PLCC} & \textbf{SRCC} & \textbf{Pb} & \textbf{Rb} & \textbf{Rg} \\
\midrule 
CLIP-Score & 0.078 & 0.082 & 0.890 & 0.369 & 0.752 \\
CLIP-IQA & 0.070 & 0.081 & 0.879 & 0.260 & 0.806 \\
NIMA & 0.221 & 0.245 & 0.700 & 0.108 & 0.744 \\
ImageReward\textsuperscript{+} & 0.271 & 0.302 & 0.905 & 0.370 & 0.908 \\
Single Reward Model\textsuperscript{+} & 0.347 & 0.362 & \textbf{0.965} & 0.387 & 0.924 \\
HFPC\textsuperscript{+} & \textbf{0.352} & \textbf{0.367} & 0.964 & \textbf{0.403} & \textbf{0.922} \\
\bottomrule 
\end{tabular}}
\end{table}

\begin{table}[ht]
\centering
\caption{Ablation study for each module in the model.}
\label{tab:ablation study}

\renewcommand{\arraystretch}{1.5} 
\setlength{\tabcolsep}{10pt} 
\resizebox{\columnwidth}{!}{ 
\begin{tabular}{@{}lccccl@{}} 
\toprule 
\textbf{Method} & \textbf{Pb} & \textbf{Rb} \\
\midrule 
ImageReward & 0.905 & 0.370 \\
Fused through features concatenation & 0.915 & 0.355 \\
Without classification loss & 0.953 & 0.342 \\
Single reward model & 0.965 & 0.387 \\
\bottomrule 
\end{tabular}}
\end{table}

\subsection{Visual Analysis}
\subsubsection{Visualization of results of HFPC}

\begin{figure}[ht]
    \centering
    \includegraphics[width=\linewidth]{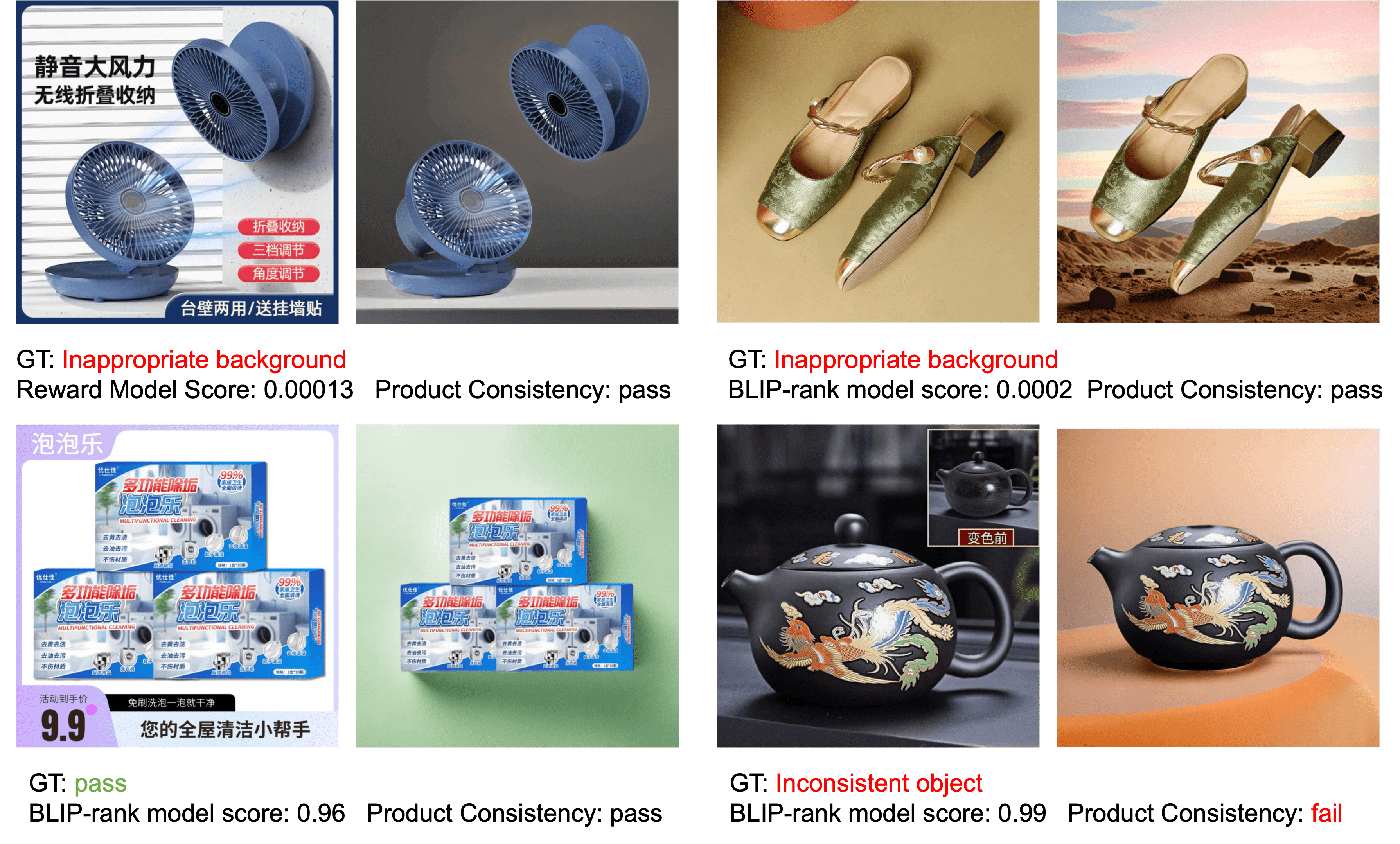}
    \caption{Visualization of results of HFPC}
    \label{fig:result}
\end{figure}

A range of results output by the HFPC are listed. Fig. \ref{fig:result} reveals that our method can accurately recognize product consistency and inappropriate background. It is worth noting that the Product Consistency Assessment module can still find low-quality generated images (i.e., fail) when the Reward-Model score is high.

\subsubsection{Visualization of Attention Mechanisms for the Reward Model}

\begin{figure}[h]
    \centering
    \includegraphics[width=\linewidth]{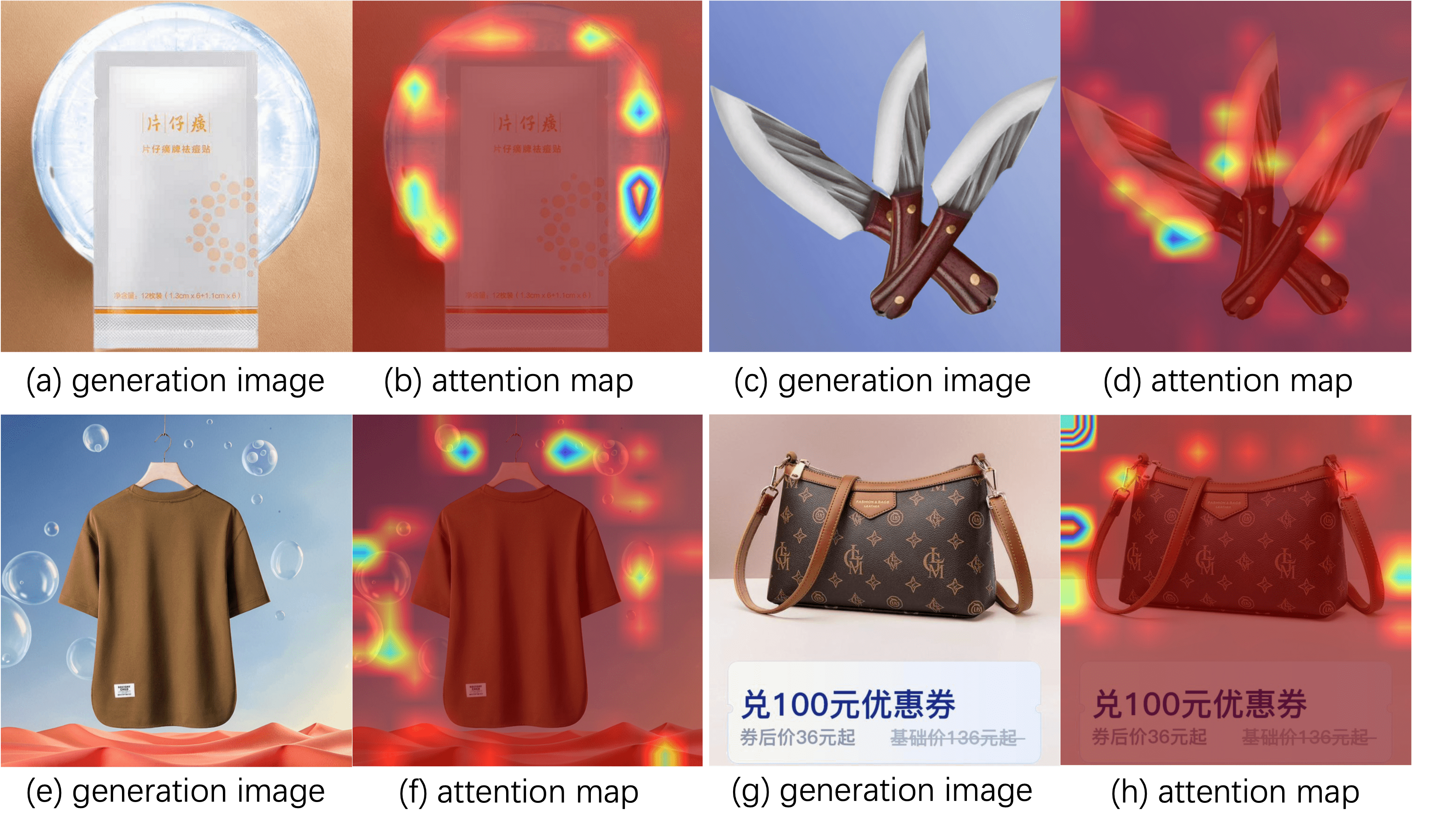}
    \caption{Attention map}
    \label{fig:cam}
\end{figure}

Our Reward Model enhances the understanding of the relationship between the original image and the generated image through an attention mechanism. This mechanism enables the model to focus on the most critical information in the image, especially the part where the products interact with the background. Thus, the attention map\citep{gao2021ts} was visualized in Fig. \ref{fig:cam} to value the model's attention in image processing.

\subsection{Inference Time and GPU Overhead}

Inference performance tests were conducted on the P100 machine (16GB), evaluating the GPU usage and inference time of both the Reward Model and Product Consistency Assessment module with batch size set to 1. Table \ref{tab:gpu_overhead} summarizes the results:

\begin{table}[htbp]
\caption{Overhead of different modules.}
\label{tab:gpu_overhead}
\centering
\begin{Huge}
    \renewcommand{\arraystretch}{1.2}
    \setlength{\tabcolsep}{10pt}
    \resizebox{\columnwidth}{!}{   
    \begin{tabular}{l|c|c}
    \toprule
    \multirow{2}{*}{Module} & \multicolumn{2}{c}{Overhead} \\
    \cmidrule(lr){2-3}
    & GPU Overhead & Average Inference Time \\
    \midrule
    Reward Model & 1508MB & 0.27s \\
    Product Consistency Assessment Module & 2566MB & 1.68s \\
    \bottomrule
    \end{tabular}}
\end{Huge}
\end{table}

\section{Conclusion}
The HFPC framework offers a novel and effective solution for the automatic quality assessment of AI-generated product images, particularly addressing the challenges of inappropriate backgrounds and product inconsistencies that are often overlooked by existing metrics. By integrating human feedback into a reward model and employing advanced segmentation techniques for precise product analysis, HFPC not only aligns more closely with human feedback but also considerably reduces the reliance on labor-intensive manual annotations. The demonstrated superiority of HFPC over other leading visual-quality-assessment models on the HFPC-44k dataset underscores its potential as a transformative tool in the realm of e-commerce and advertising. 

In the future, we will further extend the HFPC-44k dataset to include larger-scale human annotations. In addition to static data, we will introduce real-time online feedback data to train the model to incrementally accommodate real-time user preference changes ~\citep{wu2019large,castro2018end}. In addition, our proposed HFPC framework is not only capable of filtering low-quality generated images but also utilizes reinforcement learning to improve the quality of the generative models ~\citep{bai2022training,knox2011augmenting,griffith2013policy}.

\bibliography{aaai25}

\end{document}